\documentclass[11pt]{article}

\usepackage[final]{acl}

\usepackage{times}
\usepackage{latexsym}

\usepackage[T1]{fontenc}

\usepackage[utf8]{inputenc}

\usepackage{microtype}

\usepackage{inconsolata}

\usepackage{graphicx}
\usepackage{hyperref}
\usepackage{url}
\usepackage{rotating}
\usepackage{adjustbox}
\usepackage{amsmath}
\usepackage{multirow}
\usepackage{arydshln}
\usepackage{algorithm}
\usepackage{algpseudocode}
\usepackage{pifont}
\usepackage{enumitem}

\title{Syntriever: How to Train Your Retriever \hspace{-0.3em}\raisebox{-1.0ex}{\includegraphics[width=.8cm]{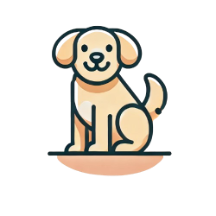}}\hspace{-0.3em} with Synthetic Data from LLMs}

\author{Minsang Kim \\
Department of Computer Science \\ and Engineering\\
Korea University \\
SK Telecom \\
South Korea \\
\texttt{kmswin1@korea.ac.kr} \\
\And
Seungjun Baek\thanks{Corresponding Author}\\
Department of Computer Science \\ and Engineering\\
Korea University \\
South Korea \\
\texttt{sjbaek@korea.ac.kr} \\
}

\begin{document}

\maketitle

\begin{abstract}
LLMs have boosted progress in many AI applications. Recently, there were attempts to distill the vast knowledge of LLMs into information retrieval systems. Those distillation methods mostly use output probabilities of LLMs which are unavailable in the latest black-box LLMs. We propose Syntriever, a training framework for retrievers using synthetic data from black-box LLMs. Syntriever consists of two stages. Firstly in the distillation stage, we synthesize relevant and plausibly irrelevant passages and augmented queries using chain-of-thoughts for the given queries. LLM is asked to self-verify the synthetic data for possible hallucinations, after which retrievers are trained with a loss designed to cluster the embeddings of relevant passages. Secondly in the alignment stage, we align the retriever with the preferences of LLMs. We propose a preference modeling called partial Plackett-Luce ranking to learn LLM preferences with regularization which prevents the model from deviating excessively from that trained in the distillation stage. Experiments show that Syntriever achieves state-of-the-art performances on benchmark datasets from various domains in nDCG@$K$. The code is available at  \href{https://github.com/kmswin1/Syntriever}{https://github.com/kmswin1/Syntriever}.
\end{abstract}

\section{Introduction}
Large Language Models~(LLMs) have become a core technology in various NLP applications such as chatbots~\cite{achiam2023gpt, team2023gemini} and coding assistants~\cite{roziere2023code, guo2024deepseek}. 
It is essential that the knowledge of LLMs is complemented by up-to-date information from external sources. To this end, retrieval-augmented generations~(RAG) have been proposed and actively explored for various knowledge-intensive NLP tasks  \cite{lewis2020retrieval, guu2020retrieval, lazaridou2022internet}. RAG enhances the LLM performance without fine-tuning by incorporating external knowledge into LLMs through search and   alleviates problems such as hallucination  ~\cite{welleck2020neural}, i.e., plausible but non-factual information generated by LLMs. 

The retrieval of documents relevant to a given query is a key task of the RAG system.  Dense retrieval methods \cite{karpukhin2020dense,gao2022unsupervised} are widely used to capture semantic relationships between queries and documents, in which text encoders are trained to learn dense embeddings of queries and passages for their semantic matching.
The encoders can be pre-trained in an unsupervised manner by using large-scale text pairs sampled from sentences and their contexts \cite{lee2019latent, izacard2021unsupervised}, and then be fine-tuned on the annotated datasets for retrieval tasks \cite{wang2022text, chen2024bge}.
Meanwhile, recent LLMs have exhibited remarkable generalization abilities in many NLP tasks, including information retrieval. 
In this paper, we explore how the vast knowledge of LLMs can be effectively utilized in training retrievers.
Recently, RePlug~\cite{shi2024replug} has been proposed for distilling the LLMs' knowledge into small retrievers. RePlug calculates the relevance scores of $k$ retrieved passages given a query, from which a likelihood over $k$ passages is computed. The retriever is trained to minimize the KL divergence between this likelihood and the LLM's likelihood over passages based on its probability of predicting the ground truth answer. However, prediction probabilities are mostly unavailable as the output in the latest \emph{black-box} 
LLMs~\cite{achiam2023gpt, team2023gemini}. Thus, we consider the distillation of LLM's knowledge into retrievers when only the synthetically generated texts are available as the output from LLMs.

\noindent\textbf{Contribution.} 
We propose Syntriever, a framework to train/fine-tune retriever models based on synthetic data so as to distill the knowledge of black-box LLMs into retrievers effectively. We propose a two-stage framework: in the first stage, called \emph{distillation stage}, we fine-tune the retriever with LLM-generated synthetic data; in the second stage, called \emph{alignment stage}, we align the retriever with the preference of LLMs. 
In the distillation stage, Syntriever exploits synthetically augmented queries using chain-of-thoughts~\cite{wei2022chain}, synthetic positive and hard-negative passages, as well as self-verification to deal with hallucination. The retriever is then trained by modified Soft Nearest-Neighbor loss \cite{frosst2019analyzing} to cluster multiple relevant passages together in the embedding space. In the alignment stage, we continually fine-tune the retriever trained from the distillation stage, where the goal is to align the retriever with  LLM preferences.
The retriever fetches top-$K$ passages from which a set of passage pairs is sampled and provided to LLMs for preference feedback. In particular, we propose a preference modeling called \emph{partial Plackett-Luce ranking} to learn LLM preferences with a regularization effect such that the aligned model does not deviate excessively from the distilled model.
We evaluated the performance of Syntriever in various domains of benchmark datasets for the retrieval tasks from BeIR~\cite{thakur2021beir}. 
Syntriever achieves superior performances on all benchmark datasets, by up to 18.6\% in nDCG@10, compared to the prior state-of-the-art. Moreover, we show that the Syntriever framework can be combined with diverse base retrievers and LLMs, leading to a significant increase in retrieval accuracy.

\begin{figure*}[t!]
    \centering
    \includegraphics[height=.7\textwidth]{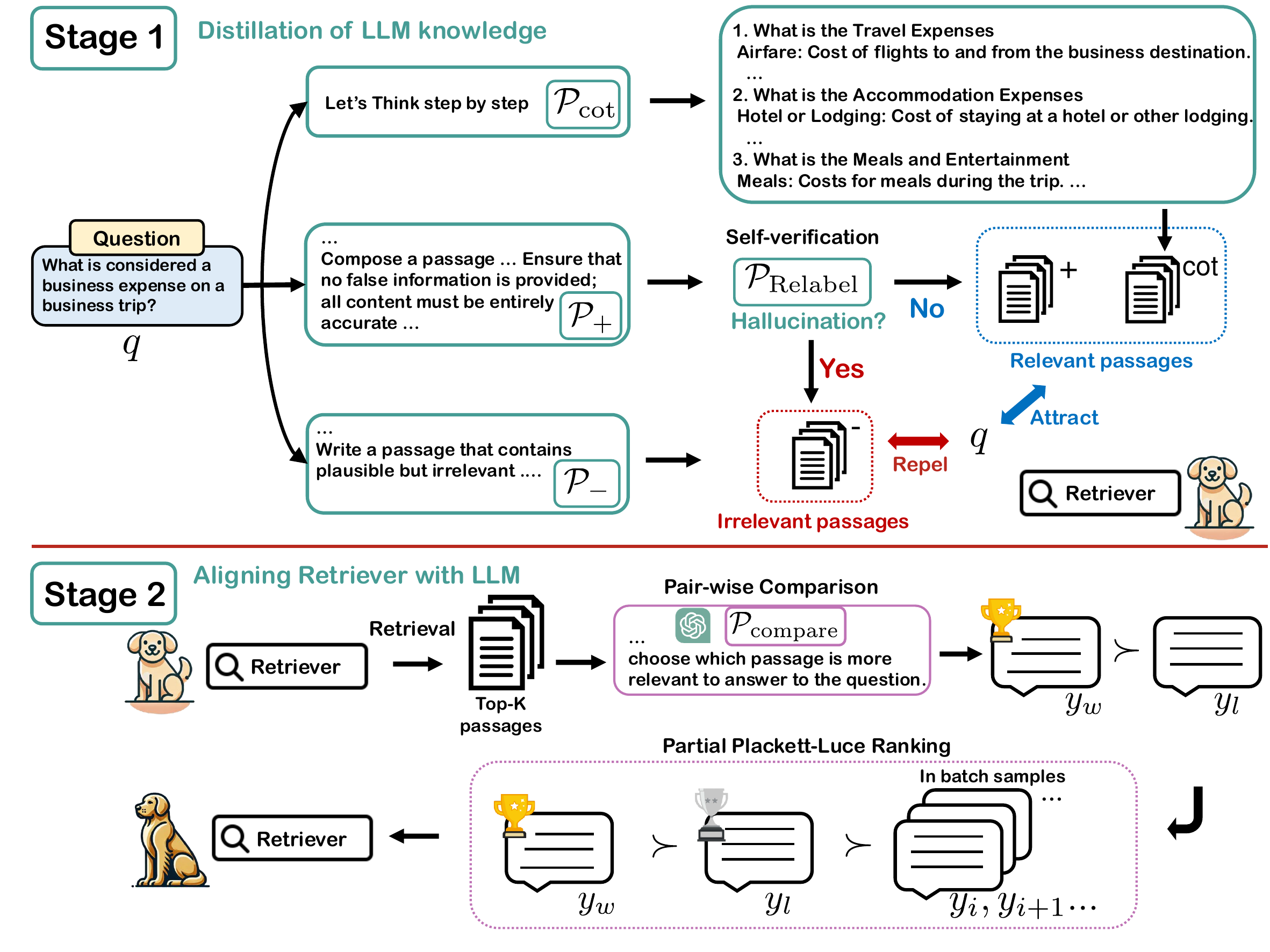}
    \caption{Overview of Syntriever. \textbf{Stage-1 (Distillation Stage).} Given a query, Syntriever uses LLMs to synthesize (i) related sub-queries (prompt $\mathcal P_\text{cot}$), (ii) relevant passages  ( $\mathcal P_+$) which are self-verified for hallucination ( $\mathcal P_\text{Relabel}$), (iii) plausibly irrelevant passages ( $\mathcal P_-$). The retriever is trained with the synthetic positive and negative passages. \textbf{Stage-2 (Alignment Stage).} The retriever is aligned with the LLM preferences. LLM compares passage pairs from top-$K$ retrieved passages. If LLM prefers $y_w$ over $y_l$, we write $y_w\succ y_l$. We propose \emph{partial Plackett-Luce ranking} to combine preference modeling and contrastive learning for the retriever to learn $y_w\succ y_l \succ$ \{in-batch negatives\}.
    }
    \label{fig:model}
\end{figure*}

\section{Training Retrievers through Passage Synthesis}

\subsection{Problem Statement and Notation}

Neural retrieval is a task of searching for top-$K$ relevant passages $\mathcal{C}$ given query $q$ using encoder $E$ from knowledge source $\mathcal{Z}$: 
\begin{equation}
    \mathcal{C} = \mathrm{Retrieval}(q, \mathcal{Z}, K, E)
\end{equation}
The retrieval system (retriever in short) is used for retrieval-augmented generations (RAG)~\cite{lewis2020retrieval,guu2020retrieval}.
Our goal is to train/fine-tune a (pre-trained) text encoder $E$ which outputs embeddings for semantic representations of queries and passages.
The semantic similarity of query $q$ and passage $p$ is measured by \[
s_\tau(q,p) := \frac{\text{sim}(E(q),E(p))}{\tau}
\]
where $\text{sim}(a,b)$ stands for the cosine similarity of vectors $a$ and $b$, and $\tau$ is the temperature hyperparameter which controls the concentration of (normalized) embeddings on the unit hypersphere.

In the training dataset, each query $q_i$ is paired with passage $p_i$ manually labeled as relevant or the answer to $q_i$. We will denote a batch of samples during training by $B$, where $B$ is a set of indices of batch samples. A typical method to train a retriever is metric learning with \emph{contrastive loss} such as InfoNCE \cite{oord2018representation, izacard2021unsupervised}:
\[
\mathcal{L}_\text{InfoNCE}=-\log\,\bigg[\frac{\exp(s_\tau(q_i,p_i))}{\sum\limits_{j\in B}\exp(s_\tau(q_i,p_j))}\bigg]
\]
That is, manually labeled $p_i$ is regarded as a \emph{positive} passage for $q_i$, and the embeddings of $q_i$ and $p_i$ are pulled closer in the embedding space. The other passages in batch $B$ are considered irrelevant to $q_i$, and as \emph{negative} passages whose embeddings are pushed away from that of $q_i$.

Next, we outline the proposed method, dubbed \emph{Syntriever}, which consists of two stages. In Stage 1 (Sec.\ \ref{stage1}), we use LLM-generated synthetic data to distill their parametric knowledge into the retriever. In Stage 2 (Sec.\ \ref{stage2}), we align the retriever with  LLM preferences. The two-stage process of Syntriever is analogous to the training of LLMs, i.e., supervised fine-tuning (SFT) followed by alignment with human preferences \cite{ouyang2022training}.  
An overview of Syntriever is depicted in Fig. \ref{fig:model}.

\subsection{Stage-1. Distillation of LLM's knowledge through Synthesis}\label{stage1}

Given query $q$, our goal is to assimilate  $q$ to a set of positive documents, and to disassimilate $q$ from negative documents. We synthesize a variety of positive and negative passages so as to distill the vast knowledge of LLMs into the retriever.

\noindent\textbf{Decomposing query to easier sub-queries.} Neural retrievers struggle with challenging queries \cite{li2024retrieval}, e.g., if a query requires {multi-step reasoning}, or is {too complex to understand}. LLMs are capable of decomposing a complex query into multiple easier sub-queries which contain fine-grained planning to answer the query. We leverage the decomposition capability by applying the original query with prompts generating chain-of-thoughts~(CoT)~\cite{wei2022chain}, e.g., ``\texttt{Let's think step-by-step}'' proposed by \cite{kojima2022large}. Specifically, given query $q_i$, we generate augmented query $q^\mathrm{cot}_i$ given by
\begin{equation}
    q^\mathrm{cot}_i = \mathcal{M}(\mathcal{P}_\mathrm{cot}(q_i))
\end{equation}
where $\mathcal{M}$($\cdot$) denotes the LLM operation, and $\mathcal{P}_\mathrm{cot}$ denotes the prompt operator to generate CoT (see Appendix \ref{appendix:prompts} for prompt details). $q^\mathrm{cot}_i$ contains sub-queries relevant to the original query, which are carefully planned out with clarification and details necessary to retrieve relevant documents. We will use $q^\mathrm{cot}_i$ as a positive document for $q_i$. This helps the retriever with understanding diverse contexts associated with related queries in the future. 

\noindent\textbf{Synthesizing positive and hard-negative passages.}  
We generate synthetic \emph{positive} and \emph{hard-negative} passages from query $q$.  Although there exist positive passages manually labeled for $q$ in the dataset, the synthesis of positive passages can distill a broader range of knowledgeable contexts from LLM to the retriever, and provide different perspectives on the query, which prevents overfitting to specific keywords or contexts. We generate synthetic positive passage $p_i^+$ related to query $q_i$ with prompt $\mathcal{P}_+$:
\begin{equation}
p^+_i = \mathcal{M}(\mathcal{P}_+(q_i))
\end{equation}

In addition, contrastive learning can be made more robust using \emph{hard-negatives}  \cite{robinson2021contrastive} where hard-negatives are samples that are difficult to distinguish from positive samples. For the retriever, hard negatives are plausible but irrelevant answers to query $q$. We synthesize hard-negative passage $p_i^-$ with prompt $\mathcal{P}_-$ given by
\begin{equation}
p^-_i = \mathcal{M}(\mathcal{P}_-(q_i))
\end{equation}

\noindent\textbf{Hallucination as Hard-negatives.} We take a step to verify whether synthetic positive passages are indeed relevant to the given query. Using LLMs to generate answers runs a risk of \emph{hallucinations}. Hallucination is a non-factual but seemingly plausible passage. The synthetic positive $p_i^+$ can potentially be a hallucination. However, \emph{the plausible irrelevance of hallucination fits the definition of hard-negatives.} Thus, we re-use hallucination as hard negative passages, which differs from prior works \cite{weng2023large,  madaan2024self} 
which simply discards hallucination outputs.  

To that end, once positive passage $p_i^+$ is synthesized, LLM checks the passage for hallucination. LLMs are known to have self-verification ability \cite{weng2023large}, i.e., they can re-verify the inferred answer. If $p_i^+$ is decided as hallucination, we label $p_i^+$ as a hard-negative, which we call \emph{Relabeling} step. Specifically,
\begin{equation}
     \hat p_i = \mathcal{M}(\mathcal{P}_{\mathrm{Relabel}}(q_i, p_i^+))
\end{equation}
where $\mathcal{P}_{\mathrm{Relabel}}$ denote the prompting for relabeling.
If $p_i^+$ is relabeled as a hard-negative,  query $q_i$ will have two hard-negatives (synthetic and relabeled) and two positives (manually labeled and CoT).

\begin{figure}[t!]
\centering
    \includegraphics[width=.5\textwidth]{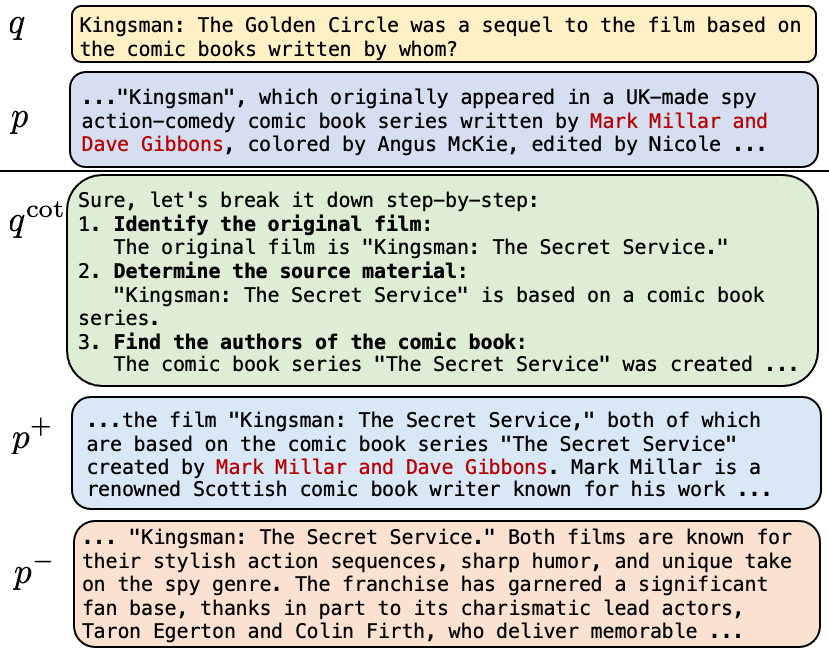}
    \caption{Example of LLM synthesis. The correct answer to the query is shown in red font.} 
    \label{fig:synthetic}
\end{figure}

In summary, given query $q_i$, positive passages are manually labeled passage $p_i$, CoT $q_i^\text{cot}$, and synthetic positive $p_i^+$ (if not relabeled as negative). The negative passages are $p_i^-$ (and relabeled passages, if any) and in-batch samples. An example of synthesized passages is shown in Fig.~\ref{fig:synthetic}.

\noindent\textbf{Putting positives together: modified Soft-Nearest Neighbor Loss.}
Next, we train the retriever with synthesized passages. Considering that there are multiple positives for a given query, we propose to use a loss inspired by soft-nearest neighbor (SNN) loss \cite{frosst2019analyzing}. SNN loss is used in metric learning for supervised classification as follows. Consider batch $B$ from a labeled dataset and a sample $x_i$ in  $B$ with label $y_i$.  The ``nearest'' neighbor (NN) to $x_i$ is selected from $B$ in a randomized fashion: the probability of $x_j$ being selected as the NN is $\propto \exp(-d_T(x_i,x_j))$ where $d_T$ is the distance metric with temperature parameter $T$. SNN loss is the negative logarithm of the probability that the NN of $x_i$ is in the same class as $x_i$:
\[
\mathcal L_{\text{SNN}}(x_i)=-\log\,\left(\dfrac{\sum_{j\in B:y_j=y_i}\exp(-d_T(x_i,x_j))}{\sum_{j\in B}\exp(-d_T(x_i,x_j))}\right)
\]
 The goal of loss $\mathcal L_{\text{SNN}}$ is \emph{entanglement} \cite{frosst2019analyzing} which is to closely cluster the sample embeddings from the same class.  

We consider a loss inspired by SNN loss. In our case, the set of points we want to cluster is a group of 4 samples ($q_i$, $p_i$, $p_i^+$, $q^\mathrm{cot}_i$). Although these groups do not represent individual classes as in SNN loss, we still want the group to be ``entangled''. Thus, similar to SNN loss, we propose a loss $\mathcal{L}_\mathrm{distill}(q_i)$ for query $q_i$ 
given by
\begin{align*}
&\mathcal{L}_\mathrm{distill}(q_i) =-\log\big(\\&  \frac{e^{s_{\tau}(q_i, p_i)}+e^{s_{\tau}(q_i, p^+_i)}+e^{s_{\tau}(q_i, q^\mathrm{cot}_i)}}
{\sum\limits_{j\in B} e^{s_{\tau}(q_i, p_j)}+e^{s_{\tau}(q_i, p^+_j)}+e^{s_{\tau}(q_i, q^\mathrm{cot}_j)}+e^{s_{\tau}(q_i, p^-_j)}}\bigg)
\end{align*}
where the similarity metric $(s_\tau)$ is used instead of the negative distance $(-d_T)$.
Another difference between $\mathcal{L}_\mathrm{distill}(q_i)$ and SNN loss is that there is no attraction term for synthetic hard-negatives ($p_i^-$) in $\mathcal{L}_\mathrm{distill}(q_i)$, i.e., they are used only for repulsion from other samples. 
\subsection{Stage-2. Retriever Alignment from LLM Feedback}\label{stage2}

Alignment is a process of aligning language models with human preferences~\cite{ouyang2022training, rafailov2024direct}. Alignment provides LMs with a pair of answer candidates for a question, where the preference between the pair is labeled by humans. We propose to align the retriever with LLM preferences as follows. Given a query, the retriever trained in the distillation stage is asked to retrieve top-$K$ passages. Next, a pair of passages is sampled from top-$K$ passages, and LLM is asked to provide the preference between the pair. Since $K$ passages are top passages from a retriever trained through the distillation stage, deciding the preference between the pair is likely to be challenging (for moderately small $K$, e.g., $K=5$). The retriever is continually trained based on the preference feedback from LLMs. The details of the alignment process are outlined as follows.

\noindent{\textbf{Step 1: Retrieve top-$K$ passages.}} Given query $q_i$, we retrieve top-$K$ passages using encoder $\hat{E}$ trained through the distillation stage:
\begin{align*}
    \hat{\mathcal{C}}_i = \mathrm{Retrieval}(q_i, \mathcal{Z}, K, \hat{E}) = \{c_{i,1}, c_{i,2}, ..., c_{i,K}\}
\end{align*}

\noindent{\textbf{Step 2: Pair-wise Comparison.}} A pair of passages, $c_{i,j}$ and $c_{i,k}$, is sampled from $\hat{\mathcal{C}}_i$. We probe LLM to decide which passage is more relevant to answer query $q_i$ using prompt $\mathcal{P}_{\mathrm{Compare}}$:
\begin{equation}    \label{eq:placket}
    (q_i,c_i^+,c_i^-)= \mathcal{M}(\mathcal{P}_{\mathrm{Compare}}(q_i, c_{i,j}, c_{i,k}))
\end{equation}
where LLM labels the more (resp. less) preferred passage as $c_i^+$ (resp. $c_i^-$). We compute the pairwise preferences of $N$ distinct passage pairs sampled from $\hat{\mathcal C}_i$ where $N\leq {K \choose 2}$ is a hyperparameter.

\noindent{\textbf{Step 3: Partial Plackett-Luce ranking.}} Consider batch $B$ of triples $(q_i, c_i^+, c_i^-)$ obtained in \textbf{Step 2} where $q_i$'s in the batch are distinct. Encoder $\hat E$ is fine-tuned with the following loss function:
 \begin{align}
        &\mathcal{L}_{\mathrm{align}}(q_i) = 
        -\log\Bigg[\dfrac{e^{s_{\tau}(q_i, c_i^+)}}{\sum\limits_{j\in B} \left(e^{s_{\tau}(q_i, c_j^+)} + e^{s_{\tau}(q_i, c_j^-)}\right)}\nonumber\\&\times\dfrac{e^{s_{\tau}(q_i, c_i^-)}}{e^{s_{\tau}(q_i, c_i^-)}+\sum\limits_{j\in B, j\neq i} \left(e^{s_{\tau}(q_i, c_j^+)} + e^{s_{\tau}(q_i, c_j^-)}\right)}\Bigg] \label{eq:ppl-loss}
    \end{align}

We refer to the training under loss (\ref{eq:ppl-loss}) as \emph{partial Plackett-Luce ranking}. The method is explained in detail as follows.  

\noindent\textbf{From Bradely-Terry to Plackett-Luce model.} Preference modeling has been used for aligning language models with human preferences~\cite{ouyang2022training, rafailov2024direct}. Bradley-Terry (BT) model \cite{bradley1952rank} is widely adopted for modeling preference over two choices.  Consider a pair of answer passages $y_w$ and $y_l$ given query $q$. If $y_w$ is preferred over $y_l$ by a human annotator, the preference relation is denoted as $y_w\succ y_l\, |\, q$. In preference modeling, it is typically assumed that there exists some (implicit) reward function $r(q,y)$ for query $q$ and answer $y$. Given query $q$ and two answers $y_1$ and $y_2$,  BT model is defined by the distribution
\begin{align}\label{eq:pair}
     p(y_1 \succ y_2 \, |\, q) = \dfrac{e^{r(q, y_1)}}{e^{r(q, y_1)} + e^{r(q, y_2)}} 
\end{align}
The fitting of BT model involves either explicitly formulating and optimizing reward $r(\cdot,\cdot)$ \cite{ziegler2019fine,ouyang2022training}, or implicitly doing so by policy optimization through parameterization \cite{rafailov2024direct}.

Plackett-Luce (PL) model~\cite{plackett1975analysis, luce1959individual} generalizes BT model to ranking $M\geq 2$ choices. Suppose $\pi:[M]\to[M]$ is a permutation. Given query $q$ and  answers $y_1,...,y_M$, we define the notation: 
\begin{align*}
p(\pi\,|\,q):=p(y_{\pi(1)}\succ y_{\pi(2)} \succ ... \succ y_{\pi(M)}\,|\,q).
\end{align*}
The PL model defines distribution $p(\pi\,|\,q)$ as
\begin{align}\label{eq:pl}
   p(\pi \, |\, q) = \prod_{m=1}^M \left(\dfrac{e^{r(q, y_{\pi(m)})}}{\sum_{j=m}^M e^{r(q,y_{\pi(j)})}}\right)
\end{align}
where the $m$-th term in the product of (\ref{eq:pl}) is the soft-max probability of the reward for the choice of rank $m$, $r(q,y_{\pi(m)})$, along with the rewards of choices of lower preferences.

\noindent\textbf{Partial Ranking through Marginalization.} The key idea of our method is to include in-batch samples in preference modeling. Consider triple $(q_i, c_i^+, c_i^-)$ from batch $B$. Our goal is to model the following preference relation:
\begin{align}
c_i^+ \succ c_i^- \succ \{\text{in-batch samples}\}\,|\, q_i\label{eq:pref}
\end{align}
where the preference ordering of in-batch samples can be arbitrary or ``don't care''. Relation (\ref{eq:pref}) is explained as follows. Firstly, $c_i^+$ is preferred over $c_i^-$ by LLM given $q_i$. Secondly, since $c_i^+$ and $c_i^-$ are in top-$K$ passages obtained from a retriever trained through the distillation stage, it is highly likely that \emph{both $c_i^+$ and $c_i^-$ are preferred over irrelevant samples in the batch.} We call this relation \emph{partial ranking}, since the ranking of the samples is incompletely specified.

The preference relation in (\ref{eq:pref}) can be modeled by \emph{marginalization} of Plackett-Luce distribution given by (\ref{eq:pl}) as follows. Suppose we want to model the preference relation
\begin{align}
y_{\pi(1)} \succ y_{\pi(2)} \succ \{y_{\pi(3)},\ldots,y_{\pi(M)}\}\,|\, q\label{eq:y}
\end{align}
where the top-two choices ($\pi(1)$ and $\pi(2)$) are preferred over the rest ($\pi(3),\ldots,\pi(M)$), and the ordering of the rest can be arbitrary.  Since $p(\pi|q)$ is a distribution over $\pi$, the distribution modeling (\ref{eq:y}) can be obtained by marginalizing $p(\pi|q)$ over the components of $\pi$ except top-two choices, $\pi(1)$ and $\pi(2)$. Specifically, we have  that
\begin{align}
&\sum_{\pi(3),\pi(4),...,\pi(M)}p(\pi \, |\, q) =\dfrac{e^{r(q, y_{\pi(1)})}}{\sum_{j=1}^M e^{r(q,y_j)}} \nonumber \\ &\times \dfrac{e^{r(q, y_{\pi(2)})}}{e^{r(q, y_{\pi(2)})}+\sum\limits_{j\neq\pi(1),\pi(2)} e^{r(q,y_j)}}\label{eq:ml}
\end{align}
The derivation of (\ref{eq:ml}) is provided in Appendix \ref{appendix:proof}.
Thus, if we set $q=q_i$, $y_{\pi(1)} = c_i^+$,  $y_{\pi(2)} = c_i^-$ and the rest of $y$'s as in-batch samples with $M=|B|$ and the reward $r(\cdot,\cdot)$ as the similarity metric $s_\tau(\cdot,\cdot)$, then (\ref{eq:ml}) models the partial relation (\ref{eq:pref}).

In conclusion, the proposed loss (\ref{eq:ppl-loss}) is the negative log-likelihood of the marginalized PL model representing partial ranking given by (\ref{eq:pref}), which makes our training a maximum likelihood estimation under distribution (\ref{eq:ml}). The key question is: \emph{why should we include in-batch samples in the preference modeling?}

\noindent\textbf{Combining Preference Modeling {and} Contrastive Learning.} Our observation is that, the training objective for preference modeling invariably takes the form of a \emph{contrastive loss}. For example, the BT model is trained with the loss which is the negative log of (\ref{eq:pair}). Suppose we use the BT model, in which case $y_1$ and $y_2$ in (\ref{eq:pair}) are replaced by $c_i^+$ and $c_i^-$ respectively. From a contrastive learning perspective, (\ref{eq:pair}) attracts $c_i^+\,(y_1)$ to $q$, but repels $c_i^-\,(y_2)$ from $q$. But this may unintentionally move $c_i^+$ and $c_i^-$ closer to samples irrelevant to $q_i$. This is undesirable, because $c_i^+$ and $c_i^-$ are among top-$K$ documents retrieved by the model trained through the distillation stage, and thus should be regarded as relatively ``positive'' and kept away from irrelevant in-batch negatives. Conventional preference modeling, such as  BT model, lacks perspective on learning with negative (irrelevant) samples.

The proposed loss directly addresses the problem: it not only captures the LLM's preferences but also maintains separation among irrelevant documents. Thus, \emph{our loss combines preference modeling and contrastive learning.} It can be seen that (\ref{eq:ppl-loss}) is simply a sum of two contrastive-type losses.
By having a similar form of contrastive loss as that from the distillation stage, e.g., positive embeddings keeping distances from in-batch negatives, our alignment loss serves as \emph{regularization}. 
That is, the model is prevented from excessively deviating from that trained in the distillation stage.
Regularization is deemed important in the alignment of LLMs as well \cite{ouyang2022training}. In addition, it is reported that the larger number of negatives leads to better performance in contrastive learning~\cite{he2020momentum}. 
In the Experiments section, we show that partial PL ranking model achieves robust performances across datasets, whereas BT model occasionally suffers from poor alignment.

\begin{table*}[t!]
\begin{adjustbox}{width=1.\textwidth,center}
\begin{tabular}{ccccccccccc}
\hline
\textbf{Method}   & BM-25 & DPR  & CoCondenser & RocketQA & Contriever & E5            & BGE-M3-EN & Nomic-Embed   & \textbf{Syntriever}             \\\hline
\textbf{MSMARCO}  & 22.8  & 36.8 & 37.4        & 30.2     & 40.3       & \textbf{42.4} & 41.3      & 37.1                     & \textbf{\underline{50.3}} \\
\textbf{HotpotQA} & 60.1  & 48.9 & 56.3        & 53.3     & 61.2       & 63.4          & 64.2      & \textbf{66.3}             & \textbf{\underline{70.2}} \\
\textbf{FiQA}     & 23.5  & 21.4 & 27.6        & 30.2     & 31.4       & 37.9          & 39.1      & \textbf{40.9}            & \textbf{\underline{41.9}} \\
\textbf{Fever}     &  \underline{\textbf{75.3}}  & 50.7    &      51.8   &   52.3   &    53.7   &   57.1  &     54.3  &    53.5  &    \textbf{60.3} \\
\textbf{SciFact}  & 66.5  & 63.3 & 48.7        & 56.8     & 64.9       & 73.7          & 75.4      & \textbf{79.1}             & \textbf{\underline{80.5}} \\
\textbf{NFCorpus} & 32.5  & 35.4 & 32.5        & 29.3     & 31.7       & 35.8          & \textbf{38.2}      & 36.8             & \textbf{\underline{43.3}} \\
\textbf{NQ}       & 32.9  & 41.2 & 43.3        & 42.1     & 42.5       & 49.3          & 46.3      & \textbf{50.1}             & \textbf{\underline{52.1}} \\\hline
\end{tabular}
\end{adjustbox}
\caption{Supervised fine-tuning results on seven BeIR benchmark with training datasets (nDCG@10). The best scores are highlighted in \underline{\textbf{bold with underline}} and, the second best scores are emphasized in \textbf{bold}.}
\label{tab:main_results}
\end{table*}

\begin{table*}[t!]
\begin{adjustbox}{width=1.\textwidth,center}
\begin{tabular}{ccccccccccc}
\hline\textbf{Method}        & BM-25 & DPR   & CoCondenser & RocketQA & Contriever & BGE-M3-EN & E5 & Nomic-Embed  & \textbf{Syntriever} \\\hline
\textbf{MSMARCO}       & 22.8  & 17.7  & 24.3       & 23.2     & 40.7       &    35.2  & \textbf{43.1}    &   26.4           &    \underline{\textbf{50.1}}              \\
\textbf{Trec-covid}    & 65.6  & 33.2  & \textbf{71.2}       & 67.5     & 59.6       & 44.6   & 61.7  & 67.1 &    \underline{\textbf{75.3}}              \\
\textbf{HotpotQA}      & 60.3  & 39.1  & 56.3       & 35.6     & 63.8       &  \textbf{68.3} & 52.4   & \textbf{\underline{69.1}}          &    60.2              \\
\textbf{FIQA}          & 23.6  & 11.2 & 27.6       & 30.2     & 32.9       & 28.3 & \textbf{37.9}   & 37.8             &   \underline{\textbf{39.5}}               \\
\textbf{Arguana}       & 31.5  & 17.5  & 29.9       & 45.1     & 44.6       & \underline{\textbf{61.5}}  & 51.4   & \textbf{54.2}          &   38.8              \\
\textbf{Touche-2020}   & \underline{\textbf{36.7}}  & 13.1  & 19.1       & 24.7     & 23.0       & 13.5 & \textbf{28.3}   & 19.0          &   19.9              \\
\textbf{Quora}         & 78.9  & 24.8  & 30.5       & 31.2     & 86.5       & \textbf{88.7 }& 87.9   &  88.4         &    \underline{\textbf{88.9}}             \\
\textbf{CQADupstack}       &   29.9  &  15.3  &    17.2    &   19.3   &  34.5   &   40.2  &  28.3   &      \underline{\textbf{49.6}}     &      \textbf{41.4}           \\
\textbf{DBPedia}       & 31.3  & 26.3  & 36.3       & 35.6     & 41.3   & 19.0    & 33.8   & \textbf{39.4}          &     \underline{\textbf{39.8}}            \\
\textbf{Climate-Fever} & 21.3  & 14.8  & 14.4       & 18.0     & \textbf{23.7}       & 18.3  & 15.4  &   \underline{\textbf{27.0}}            &     13.1            \\
\textbf{SciDocs}       & 15.8  & 7.7   & 13.7       & 13.1     & 16.5       & 9.6  & 19.0   & \textbf{19.2}         &    \underline{\textbf{19.7}}             \\
\textbf{SciFact}       & 66.5  & 31.8  & 61.5       & 56.8     & 69.3       &  71.5 &   \underline{\textbf{73.1}}   &  \textbf{71.8}           &   64.2              \\
\textbf{NFCorpus}      & 32.5  & 18.9  & 32.5       & 29.3     & 32.8       & 32.7   &  35.1 &    \textbf{35.5}             &  \underline{\textbf{36.6}}          \\
\textbf{Fever}         & \textbf{75.3}  & 56.2  & 49.5       & 67.6     & \underline{\textbf{75.8}}       & 64.3   & 58.2 & 60.3           &     60.2            \\
\textbf{NQ}            & 32.9  & 47.4  & 48.7       & 59.5     & 49.8       & 29.8  & \textbf{60.0}   & 51.2          &   \underline{\textbf{62.2}} \\\hline              
\end{tabular}
\end{adjustbox}
\caption{Zero-shot transfer results on  BeIR benchmark datasets (nDCG@10). The best scores are highlighted in \underline{\textbf{bold with underline}}, and the second best scores are emphasized in  \textbf{bold}.}
\label{tab:main_zeroshot}
\end{table*}

\section{Experiment}
\subsection{Experimental Settings}

\textbf{Datasets.} Experiments are conducted on retrieval benchmark datasets from various domains in BeIR~\cite{thakur2021beir}. We evaluate the performance of retrievers in two benchmark settings as follows.
\begin{itemize}
    \item \textbf{Supervised Fine-Tuning.} The models are evaluated on BeIR benchmark datasets which contain the training datasets. For each benchmark dataset, every model is fine-tuned on its training dataset, and we report in-domain evaluation results on that benchmark dataset.

    \item \textbf{Zero-shot Transfer.} The models are evaluated on out-of-domain datasets from BeIR benckmark. The zero-shot setting is similar to previous work \cite{izacard2021unsupervised,wang2022text}: the models can be first fine-tuned on large retrieval datasets such as MSMARCO~\cite{nguyen2016ms} and NQ~\cite{kwiatkowski2019natural} for generic knowledge, and then are evaluated on unseen datasets.
\end{itemize}
We use Normalised Discounted Cumulative Gain (nDCG@$K$) as the default performance metric.

\noindent\textbf{Baselines.} We experiment with lexical retriever BM-25~\cite{robertson2009probabilistic}, semantic retrievers DPR~\cite{karpukhin2020dense}, SBERT~\cite{reimers2019sentence}, CoCondenser~\cite{gao2022unsupervised}, RocketQA~\cite{ren2021rocketqav2}, 
Contriever~\cite{izacard2021unsupervised}, E5~\cite{wang2022text}, English model of BGE-M3-EN~\cite{chen2024bge}, Nomic-embed~\cite{nussbaum2402nomic}. 

\noindent\textbf{Settings of Syntriever.} Syntriever uses pre-trained E5 \cite{wang2022text} as the base encoder $E$. In the settings of supervised fine-tuning, Syntriever is trained with synthetic data generated from each training dataset. In the settings of the zero-shot transfer, Syntriever is first trained on synthetic data based on training datasets of MSMARCO and NQ, and then is evaluated on out-of-domain datasets from BeIR benchmarks. This is a similar setting as Contriever \cite{izacard2021unsupervised}, E5 \cite{wang2022text}, etc. To minimize the effects of model sizes on performance, we set the size of Syntriever and all baseline models to (approximately) 125M. In alignment stage of Syntriever, we set $K=5$ by default, and set $N={K\choose 2}=10$. Detailed hyperparameters are in Appendix~\ref{appendix:hyperparameters}.

\begin{table}[t!]
\begin{adjustbox}{width=.5\textwidth,center}
\begin{tabular}{|l|l|l|l|l|l|l|l|}
\hline
\textbf{$q^\mathrm{cot}$} & \textbf{$p^{+,-}$} & \textbf{$c^+ \succ c^-$} & \textbf{MSMARCO} & \textbf{HotpotQA} & \textbf{FiQA} & \textbf{SciFact} & \textbf{NFCorpus} \\ \hline
{\color{red}\ding{55}} & {\color{red}\ding{55}} & {\color{red}\ding{55}} & 42.4 & 63.4 & 37.9 & 73.7 & 35.8 \\ \hline
{\color{blue}\ding{52}} & {\color{red}\ding{55}} & {\color{red}\ding{55}} & 44.6 & 64.3 & 38.5 & 76.7 & 40.2 \\ \hline
{\color{red}\ding{55}} & {\color{blue}\ding{52}} & {\color{red}\ding{55}} & 45.7 & 64.7 & 39.3 & 77.3 & 40.8 \\ \hline
{\color{blue}\ding{52}} & {\color{blue}\ding{52}} & {\color{red}\ding{55}} & 46.2 & 65.3 & 40.2 & 78.9 & 41.4 \\ \hline
{\color{red}\ding{55}} & {\color{red}\ding{55}} & {\color{blue}\ding{52}} & 45.8 & 67.3 & 39.1 & 75.5 & 37.3 \\ \hline
{\color{blue}\ding{52}} & {\color{blue}\ding{52}} & {\color{blue}\ding{52}} & \textbf{50.3} & \textbf{70.2} & \textbf{41.9} & \textbf{80.5} & \textbf{43.3} \\ \hline
\end{tabular}
\end{adjustbox}
\caption{Ablation study (in nDCG@10). The first three columns represent the following components: synthesized query with CoT $(q^\mathrm{cot})$, synthetic positives and hard-negatives $(p^{+,-})$, alignment $(c^+ \succ c^-)$.} 
\label{tab:ablation_component}
\end{table}

\subsection{Main Results}
We first present the supervised fine-tuning results on seven datasets for the retrieval task  which are BeIR benchmarks with training datasets. The results are shown in Table~\ref{tab:main_results}.
Compared to the second-best models, Syntriever improves the retrieval performances by: 18.6\% on MSMARCO, 5.9\% on HotpotQA, 2.5\% on FiQA, 1.8\% on SciFact, 8.3\% on NFCorpus, and 4\% on NQ. The base encoder for Syntriever is a pre-trained E5; still, Syntriever achieves performance gain over E5 by: 18.6\% on MSMARCO, 10.7\% on HotpotQA, 10.6\% on FiQA, 9.2\% on SciFact, 20.9\% on NFCorpus, 5.6\% on Fever and 5.7\% on NQ. This shows that Syntriever can successfully distill LLMs' capability into small retrievers and improve their performance by a large margin. Overall, Syntriever shows robust performances on datasets both in generalized and specialized domains. Our results show that small LMs can efficiently learn from the teacher model through synthetic data and can be successfully aligned through feedback, even without access to the output probability of black-box LLMs. 

Next, we present zero-shot transfer results.
Table~\ref{tab:main_zeroshot} shows that Syntriever achieves the best performances on 8, and the second best on 1, out of 15 datasets. Note that the performances of Syntriever on MSMARCO and NQ are in-domain results, whereas other baselines, e.g., Contriever \cite{izacard2021unsupervised}, E5 \cite{wang2022text}, etc., are reported also to be trained on MSMARCO and/or NQ. In particular, although Syntriever shares the same base model as E5, it improves the retrieval accuracy on 11 datasets. This is perhaps because LLM-generated synthetic data and alignment feedback improve the generalization capabilities of the retriever on unseen data.

\subsection{Ablation study}
We conduct an ablation study on Syntriever. We add or remove model components, and the effects on the performance are shown in Table \ref{tab:ablation_component}.
The results show that both synthesized query $(q^\text{cot})$ and passages $(p^+,p^-)$ in the distillation stage improve the retrieval performances. Overall, the distillation stage achieves an average gain of 8.2\% over the base retriever. Results show that the retriever successfully learns from the parametric knowledge of LLMs during the distillation stage. Also, the alignment component $(c^+\succ c^-)$ in Table~\ref{tab:ablation_component} is shown to achieve performance gains of up to 8.8\%. Our results show that the alignment component is significant for retrieval performance, considering that nDCG@K is sensitive to the fine-grained ranking of relevant passages.

\begin{table}[t!]
\begin{adjustbox}{width=.45\textwidth,center}
\begin{tabular}{l|l|lll}
\hline
\multirow{2}{*}{\textbf{Dataset}} & \multirow{2}{*}{\textbf{Metric}} & \multicolumn{3}{c}{\textbf{Base encoder}}                                  \\ \cline{3-5} 
                                  &                                  & \multicolumn{1}{c|}{ColBERT}  & \multicolumn{1}{c|}{SBERT}        & Contriever   \\ \hline
\multirow{4}{*}{HotpotQA}         & nDCG@1                           & \multicolumn{1}{l|}{65.3 \textbf{(+10.3})} & \multicolumn{1}{c|}{63.2 \textbf{(+10.3)}}  & 73.3 \textbf{(+10.5)} \\
                                  & nDCG@3                           & \multicolumn{1}{l|}{58.2 \textbf{(+9.7)}} & \multicolumn{1}{l|}{57.0 \textbf{(+9.5)}} & 68.2 \textbf{(+9.5)}  \\
                                  & nDCG@5                           & \multicolumn{1}{l|}{60.7 \textbf{(+8.5)}} & \multicolumn{1}{l|}{59.5 \textbf{(+9.8)}} & 70.4 \textbf{(+9.7)}  \\
                                  & nDCG@10                          & \multicolumn{1}{l|}{62.8 \textbf{(+9.1)}} & \multicolumn{1}{l|}{61.8 \textbf{(+11.1)}} & 72.1 \textbf{(+10.9)} \\ \hline
\multirow{4}{*}{FiQA}             & nDCG@1                           & \multicolumn{1}{l|}{30.1 \textbf{(+3.8)}} & \multicolumn{1}{l|}{26.7 \textbf{(+5.2)}}  & 32.1 \textbf{(+5.1)}  \\ 
                                  & nDCG@3                           & \multicolumn{1}{l|}{27.8 \textbf{(+3.2)}} & \multicolumn{1}{l|}{25.1 \textbf{(+4.6)}}  & 30.4 \textbf{(+4.7})  \\
                                  & nDCG@5                           & \multicolumn{1}{l|}{30.5 \textbf{(+2.8)}} & \multicolumn{1}{l|}{26.1 \textbf{(+4.5)}}  & 31.9 \textbf{(+4.3)}  \\
                                  & nDCG@10                          & \multicolumn{1}{l|}{33.5 \textbf{(+2.9)}} & \multicolumn{1}{l|}{25.8 \textbf{(+4.3)}}  & 35.2 \textbf{(+4.6)}  \\ \hline
\end{tabular}
\end{adjustbox}
\caption{Performance gains of Syntriever with different base encoders.}
\label{tab:ablation_retriever}
\end{table}

\subsection{Performances with different encoders}
Syntriever is a framework for training encoders for retrieval, and thus can be combined with different sentence encoders. We experiment with various well-known encoders, e.g., ColBERT, SBERT, and Contriever, as the base encoders for Syntriever. Syntriever improves the performance by a large margin in all three retrieval models. The performance improvement is particularly high in nDCG@1 which concerns retrieving the exact passage relevant to the query. 
This is because the alignment stage in Syntriever helps the retriever with a fine-grained ranking of highly relevant passages. Overall, the results show that Syntriever is generally applicable to, and improves the performances of, various retrievers.

\begin{table}[t!]
\centering

\begin{adjustbox}{width=.4\textwidth}
\begin{tabular}{l|l|l|l}
\hline
\textbf{Method}  & \textbf{HotpotQA}  & \textbf{FiQA}  & \textbf{SciFact} \\ \hline
\multirow{1}{*}{w/o Self-verification} & 67.4  & 40.8 & 79.7             \\
\multirow{1}{*}{w/ Self-verification}  & 
\textbf{70.2} & \textbf{41.9} & \textbf{80.5}                 \\ \hline
\end{tabular}
\end{adjustbox}
\caption{Effectiveness of re-labeling hallucination passages. Results are in nDCG@10.}
\label{tab:ablation_verification}
\end{table}

\subsection{Effects of Re-labeling Hallucination Passages}\label{relabel}
Table~\ref{tab:ablation_verification} shows that LLM self-verification and re-labeling are effective for the synthetic training by Syntriever. The performance improvement of self-verification on HotpotQA is relatively greater than other datasets. We found that, approximately 15\% of synthetic positive passages were relabeled as hallucinations in the case of HotpotQA, whereas the proportion was about a few percent in other datasets. This indicates that the performance improvement through relabeling is likely higher for HotpotQA. In conclusion, removing hallucinations (and even \emph{re-using} them as hard-negatives as in Syntriever) through self-verification is important for data synthesis, which is the case for most tasks utilizing LLMs \cite{weng2023large}.

\begin{table}[t!]
\centering
    \begin{adjustbox}{width=.37\textwidth,center}
    \begin{tabular}{l|l|l|l}
\hline
\textbf{Dataset}          & \textbf{Metric} & \textbf{GPT-4o mini} & \textbf{GPT-4o} \\ \hline
\multirow{4}{*}{SciFact} & nDCG@1          &    65.7                                &      \textbf{66.7}                         \\
                          & nDCG@3          &   73.0                                 &        \textbf{75.0}                       \\ 
                          & nDCG@5          &   76.7                                &        \textbf{79.1}                      \\ 
                          & nDCG@10         &   76.3                                &         \textbf{78.9}                       \\ \hline
\multirow{4}{*}{NFCorpus}     & nDCG@1          &                         \textbf{46.0}      &        45.0    \\ 
                          & nDCG@3          &        \textbf{42.3}                            &        41.2    \\ 
                          & nDCG@5          &        \textbf{42.1}                            &        41.1    \\  
                          & nDCG@10         &        \textbf{42.6}                            &        41.4    \\ \hline
\end{tabular}
\end{adjustbox}
\caption{Comparison of models  trained only up to \textbf{the distillation stage} using synthetic data from GPT-4o mini vs. GPT-4o.}
\label{tab:ablation_llm_stage1}
\end{table}

\begin{table}[t!]
    \centering
    \begin{adjustbox}{width=.37\textwidth,center}
    \begin{tabular}{l|l|l|l}
\hline
\textbf{Dataset}          & \textbf{Metric} & \textbf{GPT-4o mini} & \textbf{GPT-4o} \\ \hline
\multirow{4}{*}{HotpotQA} & nDCG@1          &     64.5                               &           \textbf{68.3}                    \\
                          & nDCG@3          &   63.5                                 &    \textbf{71.2}                           \\ 
                          & nDCG@5          &     66.3                              &        \textbf{71.4}                       \\ 
                          & nDCG@10         &     68.6                              &    \textbf{70.2}                            \\ \hline
\multirow{4}{*}{FiQA}     & nDCG@1          &        41.8                            &          \textbf{42.1}                     \\ 
                          & nDCG@3          &     39.7                               &      \textbf{40.4}                         \\ 
                          & nDCG@5          &       41.3                             &          \textbf{41.5}                     \\  
                          & nDCG@10         &      40.7                              &       \textbf{41.9}                        \\ \hline
\end{tabular}
\end{adjustbox}
\caption{Comparion of models trained by preference feedback from GPT-4o mini vs. GPT-4o. Both models are trained with GPT-4o in the distillation stage.}
\label{tab:ablation_llm_stage2}
\end{table}

\subsection{Weaker but Cheaper LLMs can be effective}
We examine how the LLM capabilities affect distillation and alignment performances. We consider two LLMs: GPT-4o vs. GPT-4o-mini, where GPT-4o is the larger and more capable model. First, we compare the distillation capabilities of two LLMs. Table~\ref{tab:ablation_llm_stage1} shows the comparison, where Syntriever is trained only up to the distillation stage. Interestingly, the distillation performance of GPT-4o-mini is better than GPT-4o on NFCorpus. Considering the datasets concern different knowledge domains (SciFact: scientific, NFCorpus: medical), smaller models may be better at teaching than larger ones in certain domains. Our results interestingly coincide with recent findings that weaker models may be better at teaching than stronger models in domains like math problem solving \cite{bansal2024smaller}.
Next, we examine the alignment capabilities of LLMs. For a fair comparison, two models are first trained by GPT-4o in the distillation stage, and then trained by different LLMs in the alignment stage. Table~\ref{tab:ablation_llm_stage2} shows that the larger model (GPT-4o) is better at alignment. It is challenging to rank top-$K$ passages retrieved by a distilled retriever, requiring a deep understanding of various contexts, and thus larger models may be more favored for the task. Overall, smaller models appear to be quite competitive, i.e., the performance gap is small or even better in some domains. Thus, our prospect is that distillation/alignment through small models will become an increasingly good alternative, especially under a fixed compute budget  \cite{bansal2024smaller}.

\begin{table}[t!]
\centering
\begin{adjustbox}{width=.5\textwidth,center}
\begin{tabular}{l|l|l|l}
\hline
\textbf{Method}  & \textbf{FiQA}  & \textbf{NFCorpus} & \textbf{SciFact} \\ \hline
\multirow{1}{*}{Partial Plackett Luce}  & \textbf{41.9} & \textbf{43.3} & \textbf{80.5}\\ 
\multirow{1}{*}{Bradley Terry}  & 35.7 & 36.1  & 79.8              \\ \hline

\end{tabular}
\end{adjustbox}
\caption{Comparison of preference modeling. Results are in nDCG@10.}
\label{tab:ablation_loss}
\end{table}

\subsection{Comparison of Preference Modeling Methods}

Table~\ref{tab:ablation_loss} compares the preference modeling methods for alignment:  Bradley-Terry~(BT) \cite{bradley1952rank} and partial Plackett-Luce~(PL) ranking model. While BT and partial PL models achieve similar performances on SciFact, BT model shows poor performances on FiQA and NFCorpus. The following is a possible explanation. The search results on SciFact tend to be highly accurate, and most of top-$K$ passages are likely to contain (partly) relevant context. By contrast, top-$K$ passages on FiQA and NFCorpus which are more challenging datasets, will tend to be only marginally relevant to the given query. The partial PL performs preference ranking while keeping those marginally relevant passages away from highly irrelevant (in-batch) passages. Without such regularization of keeping marginally positive samples away from in-batch negatives, which was done during the distillation stage, BT model may cause the retriever to \emph{forget} the knowledge learned during the distillation stage. This may cause performance drops on FiQA and NFCorpus as shown in Table~\ref{tab:ablation_loss}. Thus, we conclude that the proposed partial ranking is crucial for the alignment performance. 

\begin{table}[t!]
\begin{adjustbox}{width=.5\textwidth,center}
\begin{tabular}{l|l|l|l|l|l}
\hline
\textbf{Dataset}          & \textbf{Metric}  &  
\textbf{w/o Alignment} &  $K=3$ & $K=5$ & $K=10$ \\ \hline
\multirow{4}{*}{FiQA}     & nDCG@1  &    38.7   &  
    40.9    &       42.1         &    \textbf{42.3}     \\  
                          & nDCG@3  &    38.3   &     39.7          &     40.4           &            \textbf{41.5}     \\ 
                          & nDCG@5  &    38.5   &      40.9         &             41.8   &           \textbf{42.7}      \\  
                          & nDCG@10 &    37.9   &     41.7          &             41.9   &           \textbf{42.5}      \\ \hline
\multirow{4}{*}{SciFact}     & nDCG@1  &    80.2   &      81.0         &         \textbf{81.3}       &         80.0        \\  
                          & nDCG@3  &    78.0   &     78.6          &        79.0        &        \textbf{79.8}         \\ 
                          & nDCG@5  &   78.8    &     79.7          &        \textbf{79.8}        &         79.2        \\  
                          & nDCG@10 &   78.9    &     80.0          &       80.5         &        \textbf{80.6}         \\ \hline
\end{tabular}
\end{adjustbox}
\caption{Effect of varying $K$ in  top-$K$ retrieved passages for preference alignment.}
\label{tab:ablation_topk}
\end{table}

\begin{table}[t!]
\begin{adjustbox}{width=.5\textwidth,center}
\begin{tabular}{l|l|l|l|l}
\hline
\textbf{Dataset} & Default &  $K=3$ & $K=5$ & $K=10$ \\ \hline
FiQA & 41.9 & 41.7 & 41.8 & \textbf{42.3} \\
SciFact & \textbf{80.5} & 80.0 & 80.3 & 79.8 \\
NFCorpus & 43.3 & 42.4 & 42.6 & \textbf{43.8} \\\hline
\end{tabular}
\end{adjustbox}
\caption{Effect of varying $K$ in  top-$K$ retrieved passages and the number $N$ of sampled pairs for comparison in alignment.  We set $N=K$ for this experiment. By default, Syntriever uses $K=5$ and $N=\frac{K(K-1)}{2}=10$.  The evaluation metric is nDCG@10.}
\label{tab:ablation_k_n}
\end{table}

\subsection{Effects of the number of retrieved passages during alignment} 
We examine the effect of the number $K$ in the top-$K$ passage retrieved during the alignment process. Table~\ref{tab:ablation_topk} shows the results with varying $K$, where we sample all the possible pairs, or $N={K \choose 2}$, for comparison. The performance improves with increasing $K$, up to 12\% in FiQA and 5.1\% in SciFact. Also, results show that the larger $K$, the better the performance. In addition, using a larger number of passages is particularly effective when the overall retriever accuracy is low, since it is more likely to retrieve relevant context in top-$K$-ranked passages for large $K$. However, large $K$ may incur high computational costs if $N={K \choose 2}$, and thus there is a trade-off between performance and computational overheads. In this paper, we chose $K=5$ as a good trade-off point.

In addition, we experiment with the numbers of passage pairs to be sampled for comparison ($N$) with varying $K$. Previously in Table \ref{tab:ablation_topk}, we set $N={K\choose 2}=K(K-1)/2$. Here we provide the experiments with a smaller $N$  given by $N=K$. The results are shown in   Table~\ref{tab:ablation_k_n}. Overall, if we compare Table \ref{tab:ablation_topk} and \ref{tab:ablation_k_n}, the performance seems to slightly degrade for smaller $N$. As previously, for challenging datasets such as FiQA and NFCorpus, the performance seems to gradually improve with increasing $K$, again because more retrieved passages lead to a higher chance of including relevant passages in top-$K$. At the same time, increasing $K$ seems to exhibit diminishing returns on the performance. Overall, the default setting of Syntriever $(K=5, N=10)$ appears to be a reasonable choice in terms of a balance between complexity and performance. 

\subsection{Quality of Synthetic Positives} In general, it is difficult to accurately quantify the ratio of hallucination in the synthetic passage. The passage may not have direct clues to the answers, but may contain partial information from which the answer can be deduced. How relevant a passage should be to the query so that the passage is classified as positive? This is very hard to quantify, and thus measuring the quality of synthetic passages is difficult as well.

We performed experiments to indirectly measure the quality of synthetic positives as follows. We asked GPT-4o that whether the true answer can be directly derived from synthetic positive passages (after self-verification). We asked the same question, but in this time whether the answer can be derived from the ground-truth passages provided by the dataset. The results are shown in the table below.

\begin{table}[t!]
    \centering
    \begin{adjustbox}{width=.5\textwidth,center}
    \begin{tabular}{c|c|c}
        Passages & Can derive answer& Cannot derive answer\\\hline
        Synthetic positive & 88\% & 12\% \\
        Ground truth & 84\% & 16\%
    \end{tabular}
    \end{adjustbox}
    \caption{Results of GPT-4o about whether each passage can answer ground truth. We randomly select 1000 samples in each passage set.}
    \label{tab:self_verifiacition}
\end{table}

Interestingly, GPT-4o states that only 84\% of the ground truth passages have direct clues to the true answer. This is because, a significant portion of the "ground truth" passages of the  HotpotQA dataset do not contain direct clues to the true answer, but only indirect clues or partial information. By contrast, GPT-4o stated that 88\% of synthetic positives after self-verification contain direct contexts to the true answer. Thus, we conclude that synthetic positives after self-verification are of fairly high quality.

\section{Related Work}

\paragraph{Neural Information Retrieval.}
Neural information retrieval is a key element of retrieval-augmented generation (RAG) \cite{lewis2020retrieval} which is a retrieve-and-read approach for open domain question answering tasks~\cite{chen2017reading}. 
Lexical retrieval methods such as BM-25~\cite{robertson2009probabilistic} have been mostly used prior to neural retrievals, which however had difficulties with capturing semantic information at scale. Thus, dense passage retrievers using text encoders~\cite{devlin2018bert} have been actively explored~\cite{karpukhin2020dense, gao2022unsupervised, xiong2021approximate}. 
RocketQA~\cite{qu2021rocketqa} is a multi-step training framework for a retrieval system consisting of a retriever and a re-ranker which typically is a cross-encoder to estimate the ranking among retrieved passages. RocketQA further utilizes the re-ranker to sample hard negatives from top-retrieved passages. Meanwhile, Syntriever does not use separate re-rankers, but continually trains the retriever for its alignment with the ranking preference of LLMs.
Unsupervised learning for retrieval \cite{izacard2021unsupervised, wang2022text} was proposed to train sentence encoders by contrastive learning using a large collection of text-pair datasets. Subsequently, a hybrid retrieval method which combines lexical, dense, and multi-vector retrievers has been proposed~\cite{chen2024bge}. RePlug~\cite{shi2024replug} proposed a knowledge distillation for retrievers using KL divergence associated with the prediction probabilities of relevant documents from LLMs which, however, are available only from outdated APIs. 

\paragraph{Training with Synthetic Data.}
Tiny-stories~\cite{eldan2023tinystories} first proposed training small language models using synthetic data generated by GPT-4~\cite{achiam2023gpt}. Motivated by \cite{eldan2023tinystories}, Phi~\cite{gunasekar2023textbooks} proposed filtering of code data based on the \emph{educational value} through the prompting of GPT-4.
The next version of Phi-series~\cite{li2023textbooks, abdin2024phi} generated high-quality synthetic data from judiciously selected topics in order to distill GPT-4's knowledge into small LLMs. They demonstrated that distillation through synthetic data of high educational value can boost the performances of small LLMs. \cite{wang2023improving} proposed to train a Mistral-7B model \cite{jiang2023mistral} by synthetically generating query-document pairs by prompting GPT-4 for various text embedding tasks. 
\cite{yu2024distilling} proposed distillation synthetic data where they fine-tune the student LLM using the output answers from teacher models
based on rationales~\cite{wei2022chain, deng2023rephrase}. The aforementioned methods have demonstrated that student models can efficiently learn from the synthetic data generated by teacher models. 

\section{Conclusion}
We proposed Syntriever, a training framework for retrieval systems using LLM synthesis. 
In the distillation stage, Syntriever synthesizes various types of passages including augmented queries, relevant and plausibly irrelevant passages. Relevant passages are clustered in the embedding space using modified soft nearest-neighbor loss. In the alignment stage, the retriever is continually trained based on the preference feedback of LLMs on the retrieved passages. We propose a preference modeling called partial Plackett-Luce ranking to learn LLM preferences while maintaining the similarity relation among embeddings learned during the distillation stage. Experiments show that Syntriever achieves significant performance gains over baselines on benchmark datasets from various domains.

\section{Limitations}
Although Syntriever achieves performance gains compared to baseline retrievers on various benchmark datasets, it requires LLM inferences to generate synthetic data and alignment feedback. This may incur additional costs compared to other methods which only perform a fine-tuning of text encoders. 
However, the cost of proprietary black-box LLMs has become increasingly cheaper and affordable. Moreover, weaker but cheaper LLMs become increasingly capable of teaching student models~\cite{bansal2024smaller}. Thus, we believe that the Syntriever framework is widely applicable to retrieval systems in practice.

\section{Acknowledgement}
This work was supported by the National Research Foundation of Korea (NRF) grant funded by the Korean government (MSIT) (RS-2022-NR070834), and by the Institute of Information \& Communications Technology Planning \& Evaluation (IITP)-ICT Creative Consilience Program grant funded by the Korea government (MSIT) (IITP-2025-RS-2020-II201819).

\bibliography{custom}

\appendix

\clearpage
\onecolumn
\section{Reproducibility Statement}\label{sec:reproduce}
\noindent\textbf{Source Code.} We release the source code in the \href{https://github.com/kmswin1/Syntriever}{public repository}.

\noindent\textbf{Black-box LLMs.} We experiment with \texttt{GPT-4o} and \texttt{GPT-4o-mini} for synthetic data generation. Those models are accessible by \href{https://platform.openai.com/docs/overview}{OpenAI API}. We generate LLMs' responses using our prompt templates in Appendix~\ref{appendix:prompts}.

\noindent\textbf{Training Implementation.} We implement retrieval tasks based on BEIR~\cite{thakur2021beir} which is  implemented by \href{https://sbert.net/}{sentence-transformers}. 
All the sentence encoders used in our experiments are publicly accessible in \href{https://huggingface.co/models}{HuggingFace}.

\noindent\textbf{Evaluation Datasets.} The four evaluation datasets of HotpotQA, FiQA, SciFact, and NFCorpus are released to \href{https://github.com/beir-cellar/beir}{the public repository}. 
All the experiments are conducted with the single A100 GPU with 80GB VRAM.

\section{Derivation of (\ref{eq:ml})}\label{appendix:proof}
For notational simplicity, we define \(
z_k := \exp({r(q,y_{\pi(k)})})\). We have that
\begin{align}
   \sum_{\pi(3),\ldots,\pi(M)}&p(\pi \, |\, q) = \sum_{\pi(3),\ldots,\pi(M)}\prod_{m=1}^M \left(\dfrac{z_m}{\sum_{j=m}^M z_j}\right)\label{eq:p1}
   \\&=\sum_{\pi(3),\ldots,\pi(M)}\underbrace{\dfrac{z_1}{\sum_{j=1}^M z_j}}_{(a)}\underbrace{\dfrac{z_2}{\sum_{j=2}^M z_j}}_{(b)} \prod_{m=3}^M \left(\dfrac{z_m}{\sum_{j=m}^M z_j}\right)\label{eq:p2}
   \\&=\underbrace{\dfrac{z_1}{\sum_{j=1}^M z_j}}_{(a)}\underbrace{\dfrac{z_2}{\sum_{j=2}^M z_j}}_{(b)} \underbrace{\sum_{\pi(3),\ldots,\pi(M)}\prod_{m=3}^M \left(\dfrac{z_m}{\sum_{j=m}^M z_j}\right)}_{(c)} \label{eq:p3}
   \\&=\underbrace{\dfrac{z_1}{\sum_{j=1}^M z_j}}_{(a)}\underbrace{\dfrac{z_2}{z_2 + \sum_{j\neq 1,2} z_j}}_{(b)} \label{eq:p4}
\end{align}
The derivation steps are explained as follows. \begin{itemize}
    \item 
(\ref{eq:p1}) is the definition of $p(\pi|q)$ from Placket-Luce ranking model. \item In (\ref{eq:p2}), we take two terms (a) and (b) out from the product in  (\ref{eq:p1}). \item In (\ref{eq:p3}), (a) and (b) are invariant with respect to $\pi(3),\ldots,\pi(M)$, so they can be taken out of the summation with respect to $\pi(3),\ldots,\pi(M)$. Specifically, the denominators of (a) and (b) contains the sums over $\pi(3),\ldots,\pi(M)$, and the sum is a permutation-invariant operation. Also, (c) is the sum of Placket-Luce distribution over all possible permutations of $\pi(3),\ldots,\pi(M)$. Thus, (c) must sum up to 1. \item In (\ref{eq:p4}), we simply re-write (b) in (\ref{eq:p3}) as shown in (b) in (\ref{eq:p4}). 
\end{itemize}
By replacing $z_i$ with $\exp(r(q,y_{\pi(i)}))$ in (\ref{eq:p4}), we obtain the marginalization result in Eq. (\ref{eq:ml}).

\newpage

\section{Prompt templates}\label{appendix:prompts}

\subsection{Prompt template of positive passage generation $(\mathcal P_+)$.}

\begin{figure}[h!]
    \centering
    \fbox{
    \begin{minipage}{35em}
    You are a subject matter expert in your field with substantial accumulated knowledge in a specific subject or topic, validated by academic degrees, certifications, and/or years of professional experience in that field.\\\\
    Question: \{question\}\\\\
    Write a passage that elaborates on the question. Ensure that no false information is provided; all content must be entirely accurate. Present everything you are aware of, offering a comprehensive and detailed explanation. Do not include any unverified or speculative information.
    \end{minipage}
    }
    \caption{Prompt template design for generating synthetic positive passages. }
    \label{fig:prompt_passage}
\end{figure}

\subsection{Prompts for generating plausible but irrelevant passages $(\mathcal P_-)$.}

\begin{figure}[h!]
    \centering
    \fbox{
    \begin{minipage}{35em}
    You are a subject matter expert in your field with substantial accumulated knowledge in a specific subject or topic, validated by academic degrees, certifications, and/or years of professional experience in that field.\\\\
    Question: \{question\}\\\\
    Write a passage that contains plausible but irrelevant context given the question.
    \end{minipage}
    }
    \caption{Prompt template design for generating plausible but irrelevant passages.}
    \label{fig:prompt_negative}
\end{figure}

\subsection{Prompt template of relabeling for synthetic passages $(\mathcal P_{\text{Relabel}})$.}

\begin{figure}[h!]
    \centering
    \fbox{
    \begin{minipage}{35em}
    You are a subject matter expert in your field with substantial accumulated knowledge in a specific subject or topic, validated by academic degrees, certifications, and/or years of professional experience in that field.\\\\
    Question: \{question\}\\
    Passage: \{passage\}\\
    Is the above passage relevant to the aforementioned question?\\
    Answer with yes or no.
    \end{minipage}
    }
    \caption{Prompt template design of relabeling for synthetic positive passages.}
    \label{fig:prompt_verification}
\end{figure}

\clearpage

\subsection{Prompts for pair-wise comparison of two passages $(\mathcal P_{\text{Compare}})$.}

\begin{figure}[h!]
    \centering
    \fbox{
    \begin{minipage}{35em}
    You are a subject matter expert in your field with substantial accumulated knowledge in a specific subject or topic, validated by academic degrees, certifications, and/or years of professional experience in that field.\\\\
    Passage \#1: \{passage1\}\\
    Passage \#2: \{passage2\}\\
    Question: \{question\}\\\\
    Based on your professional knowledge, choose which passage is more relevant to answer the given question.\\
    Only answer as Passage \#1 or Passage \#2
    \end{minipage}
    }
    \caption{Prompt template design for comparison of a passage pair.}
    \label{fig:prompt_rank}
\end{figure}

\section{Hyperparameters}\label{appendix:hyperparameters}
\begin{table}[h!]
    \centering
    \begin{adjustbox}{width=.6\textwidth,center}
    \begin{tabular}{|c|c|}
        \hline
        \textbf{Hyperparameter} & \textbf{Value}  \\
        \hline
        Model size & 125M \\
        Learning rate & 2e-5 \\
        Learning rate scheduler & Cosine \\
        Optimizer & Adam~\cite{kingma2014adam} \\
        Warmup ratio & 1000 steps \\
        Weight decay & 0.01 \\
        GPU Usage & Single A100 w/ 80GB VRAM \\
        Batch size & 60 (Stage 1), 100 (Stage 2) \\
        $\tau$ (temperature) & 0.05 \\
       \hline
    \end{tabular}
    \end{adjustbox}
    \caption{Detailed hyperparameters.} 
    \label{tab:hyper}
\end{table}

\section{Dataset Statistics}\label{appendix:dataset}
\begin{table}[h!]
\begin{adjustbox}{width=.7\textwidth,center}
\begin{tabular}{|l|l|l|l|l|}
\hline
\textbf{Dataset} & \textbf{Train} & \textbf{Validation} & \textbf{Test} & \textbf{Domain} \\ \hline
MSMARCO         &         532,752       &          9,261          &     7,438         &  Search Engine  \\
HotpotQA         &       170,001         &     10,895               &         14,811     &  Wikipedia  \\
FiQA             &        14,167        &     1,239                &        1,707      & Finance \\
SciFact          &         920       &        -             &       340      &    Science  \\
NFCorpus         &       110,576         &    11,386                 &       12,335    & Medical    \\
FEVER         &       140,086         &    8,080                 &       7,938    & Fact Verification    \\
NQ         &       132,803         &    -                 &       3,452    & Search Engine    \\ \hline
\end{tabular}
\end{adjustbox}
\caption{Dataset statistics.}
\label{tab:dataset_statistics}
\end{table}

\end{document}